%% file: Kor_mathematic_reasoning.tex
\pdfoutput=1

\documentclass[11pt]{article}

\usepackage{acl}

\usepackage{kotex}

\usepackage{times}
\usepackage{latexsym}
\usepackage{booktabs}       
\usepackage{amsfonts}       
\usepackage{nicefrac}
\usepackage{amsmath}
\usepackage{multirow}

\usepackage{graphicx}
\usepackage{subfigure}

\usepackage[T1]{fontenc}

\usepackage[utf8]{inputenc}

\usepackage{microtype}

\usepackage{mdframed}
\usepackage{listings}
\usepackage{xcolor}
\usepackage{longtable}
\usepackage{inconsolata}
\usepackage{microtype}

\usepackage{float}
\restylefloat{table}

\graphicspath{{image/}}
\usepackage{pdfpages}
\usepackage{url}
\usepackage{hyperref}
\usepackage{xcolor}
\usepackage{epsfig}
\usepackage{adjustbox}
\usepackage{amssymb}
\usepackage{comment}
\usepackage{caption}
\usepackage{subcaption}
\usepackage{textcomp}
\usepackage{comment}
\usepackage{relsize}
\usepackage{stmaryrd}
\usepackage{bbm}
\usepackage{rotating}
\usepackage{arydshln}
\usepackage{pifont}
\usepackage{xcolor}
\usepackage{colortbl}
\usepackage{tabularray}
\usepackage{pbox}
\usepackage[most]{tcolorbox}
\usepackage{colortbl}
\renewcommand{\lstlistingname}

\newtcolorbox{prompt}[2][]{
    colback=white,
    colframe=gray!45,
    fonttitle=\bfseries,
    coltitle=black,
    sharp corners,
    title=#2,
    #1
}
\usepackage{enumitem}
\usepackage{listings}
\newtcolorbox{instructionsbox}[1][]{
  breakable,
  colframe=cyan!75!black,    
  colback=green!5!white,     
  coltitle=black,            
  title=#1,                  
  rounded corners,           
  boxrule=0.5mm,             
  boxsep=5pt,                
  toptitle=1mm,              
  bottomtitle=1mm,           
  left=10pt,                 
  right=10pt,                
  top=5pt,                   
  bottom=5pt,                
  fonttitle=\bfseries        
}
\newtcolorbox{examples}[2][]{
    colback=white,
    colframe=green!45,
    fonttitle=\bfseries,
    coltitle=black,
    sharp corners,
    title=#2,
    #1
}
\usepackage{longtable}

\title{Do LLMs Need Inherent Reasoning Before Reinforcement Learning? \\ A Study in Korean Self-Correction}

\author{Hongjin Kim \quad Jaewook Lee \\ \textbf{Kiyoung Lee} \quad \textbf{Jong-hun Shin} \quad \textbf{Soojong Lim} \quad \textbf{Oh-Woog Kwon}
\\ ETRI \\ \texttt{\normalsize{\{drjin, benecia428, leeky, jhshin82, isj, ohwoog\}@etri.re.kr}}
}

\begin{document}
\maketitle
\begin{abstract}
Large Language Models (LLMs) demonstrate strong reasoning and self-correction abilities in high-resource languages like English, but their performance remains limited in low-resource languages such as Korean. In this study, we investigate whether reinforcement learning (RL) can enhance Korean reasoning abilities to a degree comparable to English. Our findings reveal that RL alone yields limited improvements when applied to models lacking inherent Korean reasoning capabilities. To address this, we explore several fine-tuning strategies and show that aligning the model’s internal reasoning processes with Korean inputs—particularly by tuning Korean-specific neurons in early layers—is key to unlocking RL's effectiveness. We introduce a self-correction code-switching dataset to facilitate this alignment and observe significant performance gains in both mathematical reasoning and self-correction tasks. Ultimately, we conclude that the crucial factor in multilingual reasoning enhancement is not injecting new linguistic knowledge, but effectively eliciting and aligning existing reasoning capabilities. Our study provides a new perspective on how internal translation and neuron-level tuning contribute to multilingual reasoning alignment in LLMs.

\end{abstract}

\input{sections/introduction}
\input{sections/method}

\input{sections/experiments}

\input{sections/analysis}

\input{sections/conclusion}
\input{sections/Limitations}

\input{sections/Acknowledgement}

\bibliography{anthology,custom}
\input{sections/appendix}
\bibliographystyle{acl_natbib}

\end{document}

%% file: sections/introduction.tex
\section{Introduction} \label{sec:Intro}
\begin{figure}[t]
\centering
\includegraphics[width=1\linewidth]{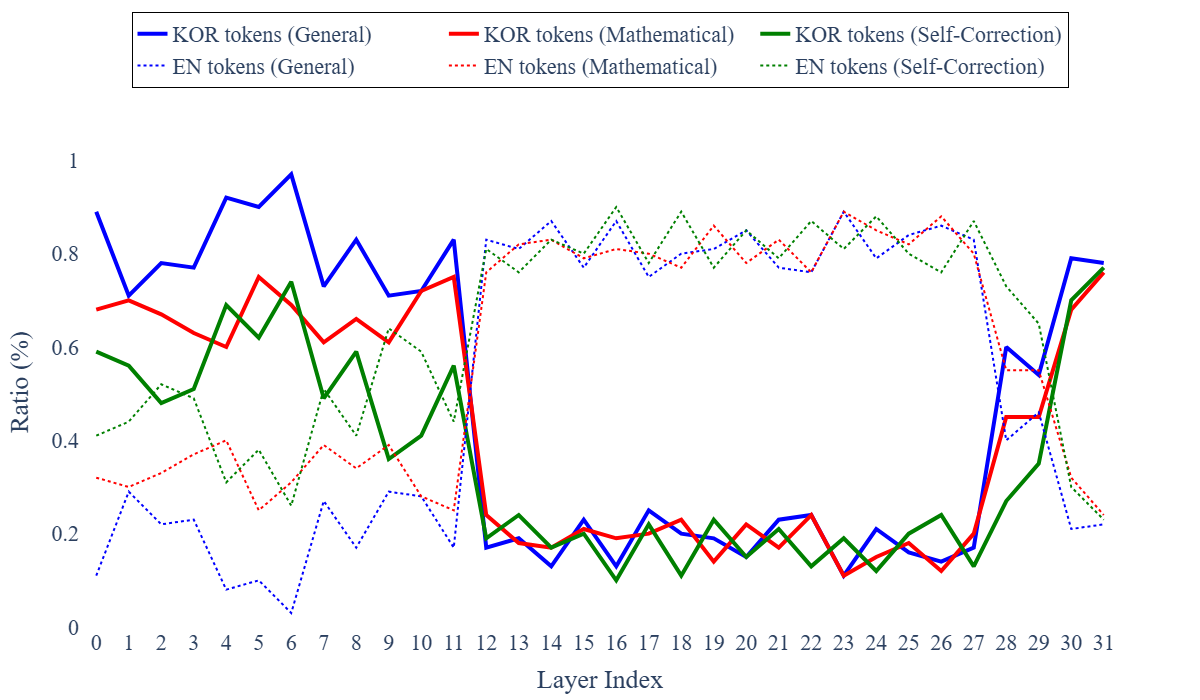}
\caption{Following \citet{zhaolarge}, we decode the hidden embeddings of an LLM (Llama3.1-8B used in this example) into the vocabulary space. In the early layers, the model internally translates Korean inputs into English. We also observe that the LLM struggles to perform this internal translation for inputs related to mathematical reasoning and self-correction.}
\label{fig:1}
\end{figure}

Large language models (LLMs) have demonstrated notable reasoning capabilities across various domains, including arithmetic, mathematics, complex problem-solving, and coding \cite{grattafiori2024llama, yang2024qwen2, abdin2025phi}. In particular, several studies have shown that training LLMs on coding tasks can enhance their mathematical reasoning abilities. Furthermore, chain-of-thought (CoT) prompting has emerged as a promising approach for effectively eliciting the reasoning capabilities of LLMs \cite{wei2022chain}. More recently, advanced techniques such as self-correction and test-time scaling via reinforcement learning (RL) have been proposed, offering further improvements in reasoning performance \cite{kumar2024training, guo2025deepseek}. However, these capabilities have primarily been observed in high-resource languages, such as English and Chinese. In the context of the synergy between coding and mathematical reasoning, it remains unclear whether non-English languages—particularly Korean, which differs significantly from English in linguistic structure—can similarly benefit from training on coding problems. Indeed, the emergence of this synergy in Korean is uncertain, given the fundamental linguistic mismatch: programming languages are inherently English-based, whereas Korean differs substantially in syntax, vocabulary, and writing system. The insufficiency and imbalanced distribution of language resources in pre-training data further contribute to discrepancies in the reasoning capabilities of LLMs across languages. To mitigate this gap, researchers have explored transferring reasoning capabilities from high-resource languages to low-resource ones \cite{shen-etal-2024-language, yoo2024code, ko2025understand}. Although \citet{yoo2024code} and \citet{ko2025understand} report improvements in LLM performance on Korean tasks, their work lacks interpretability regarding how LLMs handle Korean input and reasoning. Specifically, \citet{yoo2024code} proposed a pre-training method based on curriculum code-switching, while \citet{ko2025understand} introduced a pipeline in which the model receives and generates in English, and then translates the output into Korean. \\ \indent
To address this lack of interpretability, this study conducts an empirical and comprehensive investigation into the internal behavior of LLMs when they process Korean mathematical and self-correction reasoning. Unlike prior works that primarily highlight performance gaps between English and Korean on the same-task datasets, we assess the generation difficulty of LLMs in these two languages. We observe that LLMs face greater challenges with mathematical reasoning and self-correction tasks in Korean, compared to general tasks such as question answering and knowledge extraction. This discrepancy results in notable performance degradation on Korean mathematical reasoning benchmarks. Furthermore, inspired by the findings of \citet{zhaolarge}—which suggest that LLMs tend to internally translate multilingual inputs into English (or another dominant pre-training language), reason in that language, and then generate outputs in the original input language—we examine the behavior of LLMs across layers. Figure~\ref{fig:1} illustrates the ratio of Korean and high-resource language tokens across layers when processing Korean inputs.
Inspired by the results in Figure \ref{fig:1}, we hypothesize that LLMs may experience greater difficulty \textbf{\textit{internally translating}} Korean into English in tasks involving mathematical reasoning and self-correction than in general tasks.
Based on these assumptions, we raise the following research question: \textbf{Can LLMs enhance their Korean reasoning abilities through RL and achieve benefits comparable to those observed in English?} This question is motivated by recent findings suggesting that RL can effectively enhance reasoning abilities in LLMs when the models already possess sufficient underlying reasoning capabilities \cite{liu2025understanding}. In other words, we investigate whether LLMs can improve their Korean reasoning abilities via RL even in the absence of strong inherent Korean reasoning capabilities. \textbf{Our results indicate that RL alone offers limited benefit for enhancing Korean reasoning performance compared to English.} This leads us to further ask: \textbf{If LLMs already possess sufficient inherent Korean reasoning capabilities prior to RL, does RL yield greater improvements?} We find that, indeed, \textbf{RL is significantly more effective when the model has already acquired a strong foundation in Korean reasoning.} In particular, for self-correction tasks, this process helps LLMs better reflect and utilize their internal reasoning processes. In this study, we also examine effective fine-tuning strategies for enhancing Korean reasoning capabilities, as this is crucial for improving the effectiveness of RL. \textbf{Ultimately, we conclude that the key to success lies not in injecting Korean reasoning abilities into LLMs, but in effectively eliciting their existing English reasoning capabilities and aligning them with Korean inputs.}

%% file: sections/method.tex
\section{Method}
In this section, we describe the methodology used to assess the generation difficulty of LLMs on mathematical reasoning and self-correction tasks in Korean, in comparison to general tasks. To enable this analysis, we construct a Korean self-correction dataset using the MathDial corpus and detail the data collection process. We also briefly outline the fine-tuning methods employed to evaluate their effectiveness in enhancing the benefits of RL. Finally, we provide a concise overview of the RL approach used in our experiments.
\subsection{Measuring Generation Difficulty} \label{sec:CAS_and_DAS}
\citet{li-etal-2024-quantity} introduced the Conditioned Answer Score (CAS) and Direct Answer Score (DAS) metrics to identify discrepancies between a model's expected responses and its intrinsic generation capability. We adopt CAS and DAS to measure generation difficulty across general tasks, mathematical reasoning, and self-correction outputs. CAS is designed to assess a model’s ability to generate a target response given an instruction, and is defined as follows:
\begin{equation} \label{eq:CAS}
\centering
\small
    s_{\theta}(R|I) = -\frac{1}{T}\sum^T_{i=1}logP(w^R_i|I, w^R_1, w^R_2, ..., w^R_{i-1};\theta)
\end{equation}
Here, $T$ is the number of tokens in the target response $R$. In our study, the instruction $I$ may include a CoT prompt, a self-correction prompt, a math problem, or a general query. As shown in Equation~\ref{eq:CAS}, CAS is computed as the average cross-entropy loss and serves a similar role to perplexity, since exponentiating CAS yields the perplexity of generating $R$ given $I$. This metric captures how well the model’s response aligns with both the instruction and the corresponding correct answer. However, a high value of $s_{\theta}(R|I)$ does not necessarily imply that the instruction is difficult to follow; it may instead reflect the inherent complexity of the target string $R$. Therefore, we also employ the DAS to evaluate the model's ability to generate the response in isolation:
\begin{equation} \label{eq:DAS}
\centering
\small
    s_{\theta}(R) = -\frac{1}{T}\sum^T_{i=1}logP(w^R_i|w^R_1, w^R_2, ..., w^R_{i-1};\theta)
\end{equation} 
\noindent
DAS measures the intrinsic difficulty of generating $R$ without any instruction. A higher DAS indicates that the target response is inherently more challenging or complex for the model to produce.
Using CAS and DAS, we can measure the generation difficulty of general tasks and mathematical reasoning problems based on widely used existing datasets.
\subsection{Self-correction Dataset}
\begin{figure}[t]
    \centering
    \includegraphics[width=1.0\linewidth]{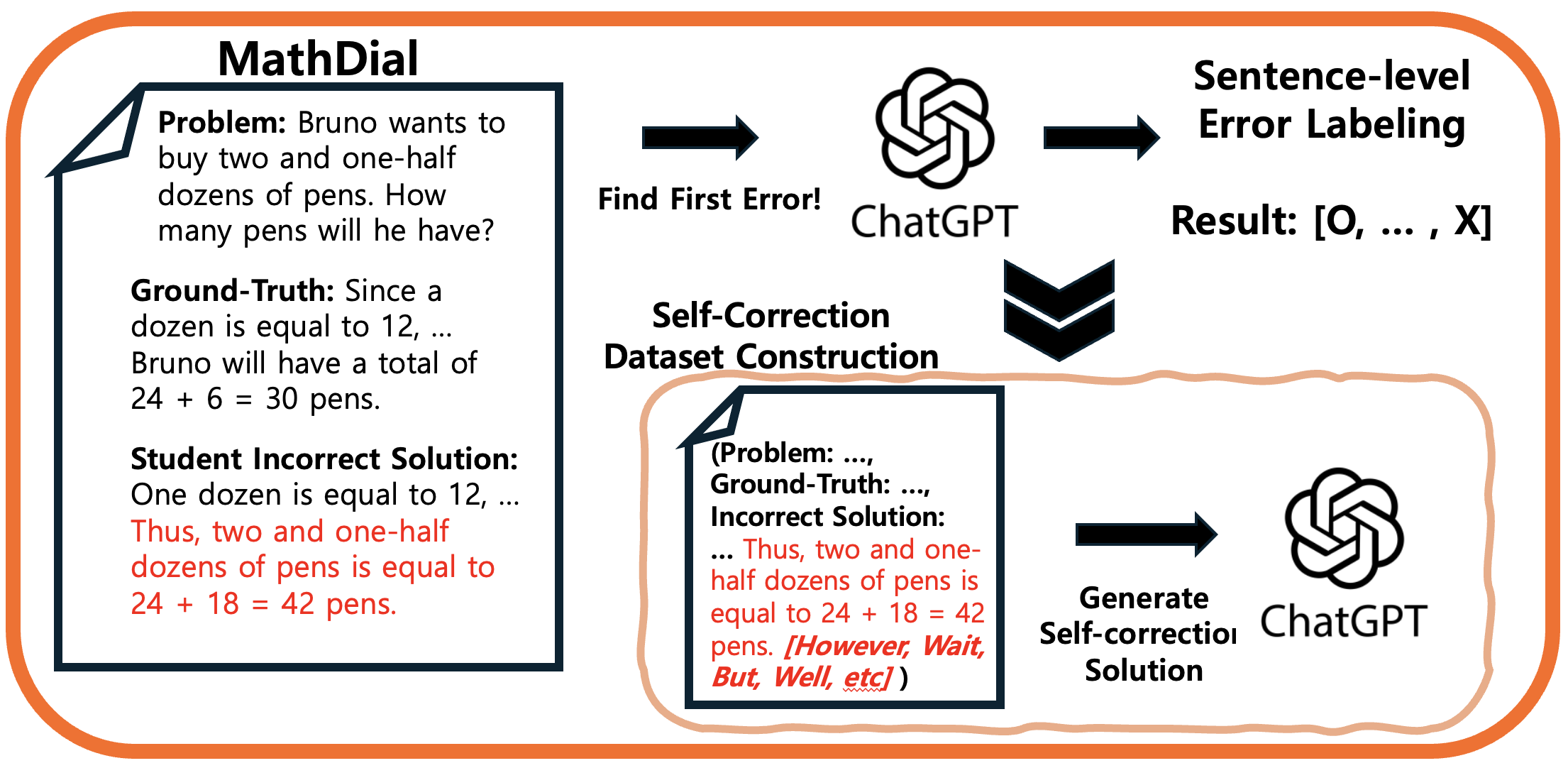}
    \caption{Overall Process of Self-Correction Dataset Construction.}
    \label{fig:self-correction}
\end{figure}
There is a lack of datasets specifically designed to evaluate self-correction capabilities. To address this, we utilize the MathDial dataset \cite{macina2023mathdial}, which includes math problems along with students’ incorrect solutions. For our evaluation, we require not only the problem and incorrect solution but also a corresponding self-corrected solution—one that initially contains errors but ultimately arrives at the correct answer through self-reflection.
To construct such self-corrected solutions, we provide GPT-4o \cite{hurst2024gpt4o} and GPT-4.1 with a math problem and a student’s incorrect solution from the MathDial dataset. We then instruct the model to (1) locate the first error in the solution and (2) starting from that point, generate a revised self-corrected solution that identifies and fixes the mistake en route to the correct answer (Figure \ref{fig:self-correction}).
To evaluate the generation difficulty of self-correction, we input the following into the target LLMs: a self-correction instruction, the math problem, and the student's incorrect solution up to the first error. We then concatenate this input with self-correction trigger words (e.g., "however", "wait", etc.) to prompt the LLMs to generate a self-corrected response. This allows us to assess the model's ability to revise its own reasoning based on minimal guidance. This process is first conducted in English, the original language of the MathDial dataset. We then translate the dataset into Korean and repeat the same procedure to obtain parallel Korean self-correction samples.
The detailed prompts and examples of generated self-correction responses are provided in Appendix~\ref{app:self-correction_details}.

\subsection{Fine-tuning Approach} \label{sec:fine-tuning}
\paragraph{Continual Pre-training:} We further pre-train LLMs on the self-correction code-switching dataset to enhance Korean mathematical reasoning capabilities and to enable inherent self-correction abilities in Korean, following the approach of \citet{yoo2024code}.
\paragraph{Specific Layer Tuning:} Following the knowledge editing technique \cite{wang2024detoxifying}, we identify the most critical layer and fine-tune it to improve mathematical reasoning and self-correction performance in Korean.
\paragraph{Adapter Tuning:} We apply Low-Rank Adaptation (LoRA) \cite{hu2022lora} to efficiently fine-tune the LLMs for enhancing mathematical reasoning and self-correction in Korean.
\paragraph{DPO (Direct Preference Optimization):} We adopt DPO \cite{rafailov2023direct} to guide the LLMs toward preferring self-corrected responses over the original ground-truth answers in the MathDial dataset.

\paragraph{Neuron Identifying}
Unlike Language Activation Probability Entropy (LAPE), which identifies neurons only in FFN modules \cite{tang-etal-2024-language}, we follow Parallel Language-Specific Neuron Detection (PLND) \cite{zhaolarge}, which detects language-specific neurons in both self-attention and FFN layers.
To detect language-specific neurons, we must assess the significance of a given neuron with respect to a specific input. Let $h_i$ denote the hidden representation before the $i$-th layer of a Transformer model when processing input $c$, and let $h_{i+1} = M_i(h_i)$ represent the hidden representation after the $i$-th layer, where $M_i$ denotes the parameters at that layer.
For a particular neuron in the $i$-th layer, denoted as $N^{(i)}$—whether located in the self-attention or FFN submodule—we quantify its importance for processing input $c$ by measuring the difference in the output $h_{i+1}$ when the neuron is activated versus deactivated. Formally, the impact of neuron $N^{(i)}$ on input $c$ is defined as:
\begin{equation}
\centering
\small
\text{Imp}(N^{(i)}|c) = \left|M_i \backslash N^{(i)}(h_i) - M_i(h_i)\right|_2
\end{equation}
Here, $M_i \backslash N^{(i)}(\cdot)$ refers to deactivating neuron $N^{(i)}$ within $M_i$ by setting all of its parameters to zero.
Given a set of $n$ input sequences in a specific language, denoted as $C = \{c_1, ..., c_l, ..., c_n\}$, we compute the importance of each neuron in each layer for each input. We select neurons whose importance scores fell within the top 1\%, following prior work \cite{tang-etal-2024-language}, across all inputs in $C$:
\begin{equation}
\centering
\small
\{N^{(i)} \mid \text{Imp}(N^{(i)}|c_l) \geq \epsilon, \ \forall c_l \in C \}
\end{equation}
This sequential neuron detection process requires traversing all neurons and inputs, making it computationally expensive. A more detailed description of a parallelized algorithm designed to accelerate this process \cite{zhaolarge} and the number of languages used are provided in Appendix~\ref{app:identifying_neurons}.

\paragraph{Fine-tuning Language-Specific Neurons}
In this study, we further fine-tune the identified language-specific neurons using the following training objective:
\begin{equation}
\centering
\small
\mathcal{L} = -\sum_{t=1}^T \log \, p(x_t \mid x_{<t}; \theta_{\text{frozen}} + \Delta_S)
\end{equation}
Here, $S$ denotes the set of language-specific neurons selected based on their importance.

Our goal is to align LLMs by enhancing internal translation through the tuning of language-specific neurons. To effectively achieve this, we construct a self-correction code-switching dataset in which the solution reasoning progresses through three stages: initially in English only, then in a mixture of English and Korean, and finally in Korean only. It is important to note that the model was not trained in three separate stages (English, then mixed, then Korean). Instead, the dataset itself is structured so that each reasoning trace comprises a sequence of outputs—first in English, followed by mixed English–Korean, and finally in Korean. The model is trained on this unified dataset in a single pass, learning to generate self-correction traces that progressively transition across languages. To generate these self-correction code-switching solutions, we use the MathDial dataset and prompt GPT-4o and GPT-4.1 to generate solutions given a math problem and its corresponding gold-standard solution. The detailed prompts and examples of generated self-correction code-switching responses are provided in Appendix~\ref{app:self-correction_details}.

\subsection{Reinforcement Learning}
\paragraph{GRPO} In this study, we adopt Group Relative Policy Optimization (GRPO) as our RL methodology \cite{guo2025deepseek}. GRPO improves computational efficiency by removing the need to learn a complex value function—typically required in conventional methods such as PPO—whose size is comparable to the policy model. Instead, GRPO estimates the advantage by generating multiple response candidates for a given input and then computing relative rewards by comparing and normalizing each response against others within the same group. This group-relative advantage is directly used to update the policy model, thereby reducing the computational burden and mitigating the learning instability often encountered in RL training of LLMs.
To assign rewards, we adopt two strategies. The first is an \textit{outcome-based reward}, which depends on the correctness of the final response: a score of +2 is given for a correct answer and -2 for an incorrect one. The second is a \textit{format-based reward}, designed to encourage the model to explicitly present its reasoning using tags such as \textsc{<think>...</think>}. In this case, a reward of +1 is assigned if the model adheres to the specified format, and -1 otherwise.

%% file: sections/experiments.tex
\section{Experiments}


\subsection{Experimental Setup}
We employed GPT-4o and GPT-4.1 to generate code-switching self-correction data. Additionally, we used LLaMA 3.1 8B \cite{llama3modelcard} and Qwen 2.5 \cite{qwen2} 7B and 32B models to detect and fine-tune language-specific neurons for our experiments. For reasoning optimized models, we used Qwen3 \cite{yang2025qwen3} 8B. Detailed implementation information is provided in Appendix~\ref{app:experimental_setting_details}.
\paragraph{Datasets}
For fine-tuning language-specific neurons, we used the self-correction code-switching dataset generated by GPT-4o and GPT-4.1. To evaluate the generation difficulty and demonstrate the effectiveness of our neuron tuning method for enhancing mathematical reasoning and self-correction capabilities, we utilized the MathDial \cite{macina2023mathdial} and HRM8K \cite{ko2025understand} datasets. HRM8K\footnote{https://huggingface.co/datasets/HAERAE-HUB/HRM8K} is a bilingual math benchmark (including GSM8K \cite{cobbe2021training}, MATH \cite{hendrycks2021measuring}, and Omni-MATH \cite{gao2024omni}) consisting of both Korean and English, making it suitable for evaluating mathematical reasoning performance in both languages. Additionally, to verify whether our neuron fine-tuning approach affects general language understanding and question answering performance in English, we employed the MMLU \cite{hendrycks2020measuring} and GPQA \cite{rein2024gpqa} datasets. A statistic of datasets is in Appendix \ref{app:statistics}
\paragraph{Baselines}
To investigate which training technique most effectively enhances Korean reasoning capabilities—and thereby improves the effectiveness of RL—we adopt several approaches, as described in Section \ref{sec:fine-tuning}. We train the LLMs using these techniques with our self-correction code-switching dataset (Example of data is provided in Appendix \ref{app:self-correction_details}).

\subsection{Experimental Results}
\begin{figure}[t]
    \centering
    \includegraphics[width=1.0\linewidth]{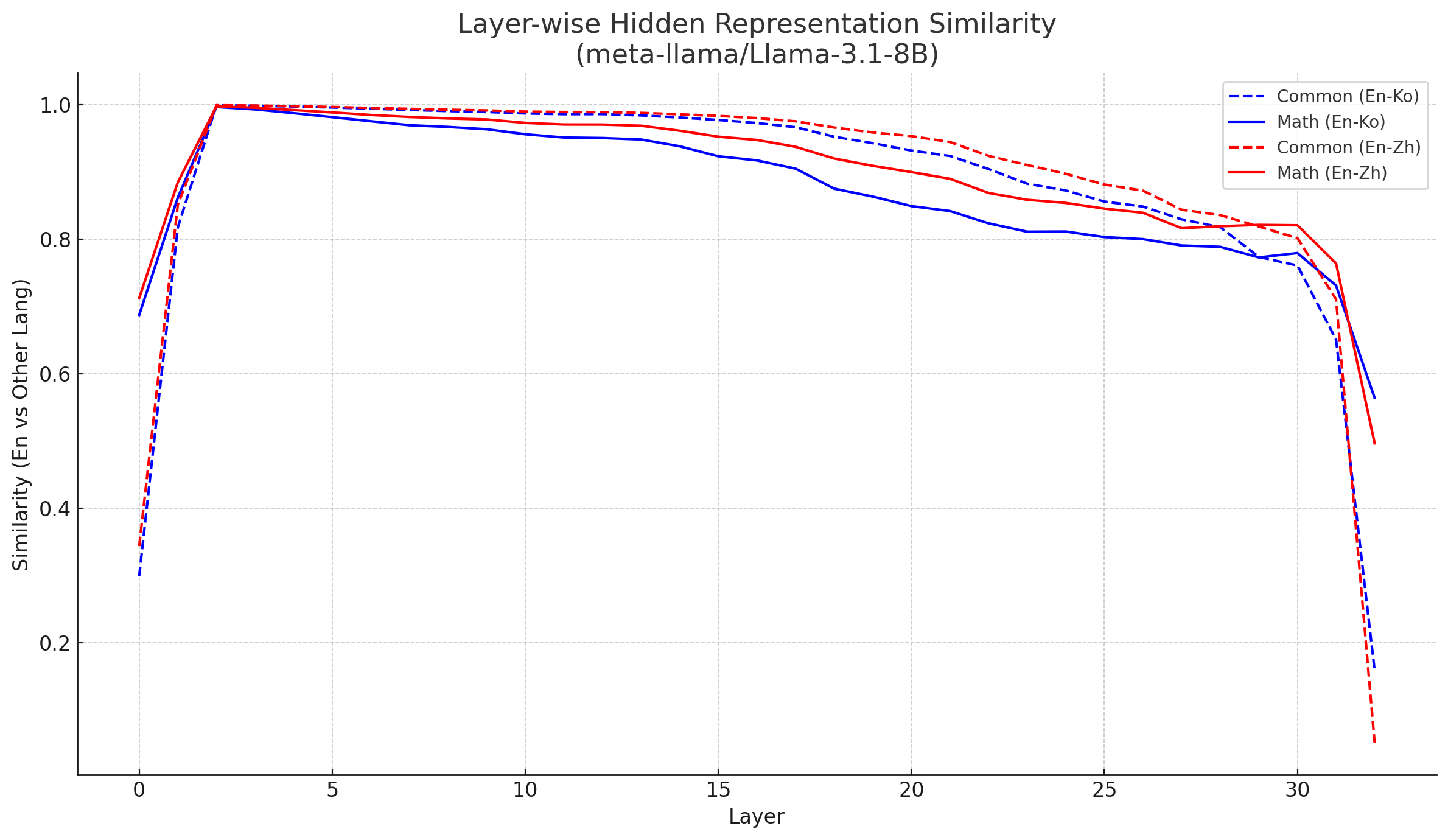}
    \caption{Layer-wise comparison of hidden representations for English vs. Chinese and Korean inputs on general QA and MATH datasets.}
    \label{fig:hidden_representation}
\end{figure}
To substantiate our assumption that internal translation (or “understanding”) begins in the early layers, we compare the hidden representation similarities among English, Korean, and Chinese inputs. As shown in Figure~\ref{fig:hidden_representation}, the representations of Korean and Chinese become highly similar to those of English even in the early layers, with similarity scores approaching. These findings suggest that the model begins mapping non-English inputs onto English representations early in the network, after which reasoning proceeds primarily in English in the middle layers. Because common sentences are well aligned, they consistently maintain very high similarity scores. Furthermore, since Chinese has been trained with substantially more tokens than Korean, its hidden representations exhibit even higher similarity with English, suggesting that internal translation occurs more effectively. We therefore believe that stronger internal translation implies better understanding and problem-solving ability on the given tasks.
\subsubsection{Results of RL without Sufficient Reasoning Ability} \label{sec:RL_before_tuning}
\begin{table*}[t]
\centering
\small
\begin{tabular}{llcc}
\toprule
Model                        & Dataset     & Accuracy ($\Delta$) After RL & Actual Self-Correction Behavior \\
\midrule
\multirow{2}{*}{Llama3.1-8B} & GSM8K (KOR) & +1.1                          & 4\%                             \\
                             & MATH (KOR)  & +0.2                          & 6\%                             \\ \midrule
\multirow{2}{*}{Qwen2.5-7B}  & GSM8K (KOR) & +1.4                          & 7\%                             \\
                             & MATH (KOR)  & +0.2                          & 10\%                            \\ \midrule
\multirow{2}{*}{Qwen3-8B}  & GSM8K (KOR) & +2.7                          & 2\%                             \\
                             & MATH (KOR)  & +3.4                          & 29\%                            \\
\bottomrule
\end{tabular}
\caption{Performance of RL without sufficient Korean reasoning ability. Accuracy improvement ($\Delta$) and observed actual self-correction behavior are reported for Llama and Qwen models on GSM8K and MATH datasets in Korean.} \label{tab:before_train}
\end{table*}

\citet{kumar2024training} identified two major failure modes in SFT-based self-correction training: (1) distribution shift, where models can correct errors from the base model but fail to correct their own, and (2) behavior collapse, where models produce strong initial responses with little or no revision. Motivated by these findings, we further analyze the self-correction behavior of LLMs in Korean. Specifically, we apply RL to vanilla LLMs and assess whether the RL-augmented models demonstrate genuine self-correcting capabilities.
For this experiment, we adopt GRPO \cite{guo2025deepseek}; detailed training settings are provided in Appendix~\ref{app:experimental_setting_details}. We evaluate self-correction performance on GSM8K and MATH by selecting 100 samples from each dataset where the model initially fails and produces incorrect solutions. These incorrect solutions, along with their corresponding problems, are then provided as input to the RL-trained models. We evaluate whether the models successfully identify and correct their initial mistakes to arrive at the correct answers. Table~\ref{tab:before_train} presents the performance of RL applied to vanilla LLMs. As shown, applying RL without sufficient Korean reasoning ability neither elicits inherent self-correction behavior in Korean nor effectively enhances Korean reasoning capabilities. To investigate the underlying cause of this phenomenon, we first measure the generation difficulty of self-correction outputs in Korean.

\subsubsection{Results of Generation Difficulty}
\begin{table*}[t]
\centering
\small
\begin{tabular}{lcccccccc}
\toprule
\multirow{2}{*}{Model}       & \multirow{2}{*}{Language} & \multirow{2}{*}{Metric $\downarrow$} & \multicolumn{6}{c}{Dataset}                              \\ \cline{4-9}
                             &                           &                         & MMLU & GPQA & GSM8K & MATH & Omni-MATH & Self-Correction \\ \midrule
\multirow{4}{*}{Llama3.1-8B} & \multirow{2}{*}{EN}       & CAS                     &  2.2    &  2.3    &  2.1     &  2.3    &   2.9        &  2.7               \\
                             &                           & DAS                     &  0.9   &   1.1   &   2.8    &   3.1   &    3.6       &   3.9              \\ \cline{2-9}
                             & \multirow{2}{*}{KOR}      & CAS                     &  2.5    &  2.2    &  \textbf{4.6}     &  \textbf{5.1}   &  \textbf{5.0}         &   \textbf{5.8}              \\
                             &                           & DAS                     &  1.3    &  1.4    &  \textbf{4.7}     &  \textbf{5.5}  &   \textbf{5.7}       &    \textbf{5.9}             \\ \midrule
\multirow{4}{*}{Qwen2.5-7B}  & \multirow{2}{*}{EN}       & CAS                     &  1.4    &  1.7    &  1.8     &  2.0    &   2.5        &  2.6               \\
                             &                           & DAS                     &  0.5   &   0.8   &   2.5    &   2.7   &    3.2       &   3.8              \\ \cline{2-9}
                             & \multirow{2}{*}{KOR}      & CAS                     &  1.6    &  1.6    &  \textbf{4.0}     &  \textbf{4.3}   &  \textbf{4.0}         &   \textbf{5.6}              \\
                             &                           & DAS                     &  0.6    &  0.9    &  \textbf{4.4}     &  \textbf{4.8}  &   \textbf{4.6}        &    \textbf{5.6} \\ \bottomrule
\end{tabular}
\caption{Results showing the generation difficulty differences between general tasks and mathematical reasoning/self-correction tasks. Experiments are conducted in a zero-shot setting, with LLMs prompted using three instruction types: vanilla, CoT, and self-correction. CAS is assessed using these instructions, while DAS is measured without instructions for mathematical reasoning and self-correction tasks. For multiple-choice question answering outputs, we convert them into generative formats when evaluating DAS, as this metric assesses the difficulty of generating the answer alone. The reported performance for each metric is averaged across the three instruction types.}
\label{tab:results_CAS_DAS}
\end{table*}

\begin{figure}[t]
    \centering
    \includegraphics[width=\linewidth]{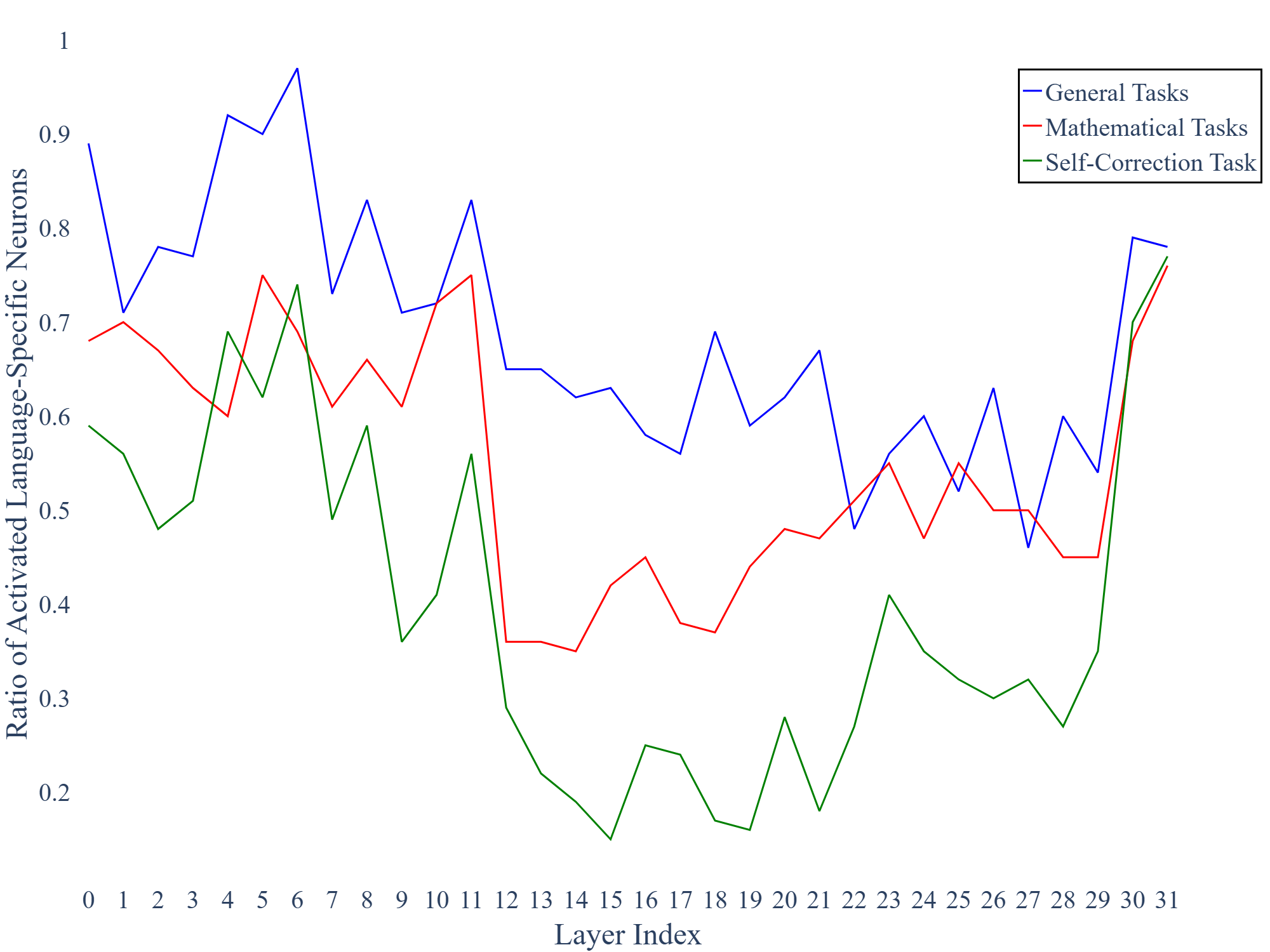}
    \caption{Ratio of activated Korean-specific neurons across tasks (Llama3.1-8B model).}
    \label{fig:ratio_activated_Koran}
\end{figure}

\begin{table}[t]
\centering
\small
\resizebox{\linewidth}{!}{
\begin{tabular}{lcc}
\toprule
\multirow{2}{*}{Model}       & \multirow{2}{*}{Language} & Dataset                       \\ \cline{3-3}
                             &                           & Self-Correction (exact-match) \\ \midrule
\multirow{9}{*}{Llama3.1-8B} & \multirow{1}{*}{EN}       & \multirow{1}{*}{\textbf{67.84}}        \\ \cline{2-3}
                             & \multirow{1}{*}{EN Ins.}      & \multirow{1}{*}{\underline{40.81}}        \\ \cline{2-3}
                             & \multirow{1}{*}{KOR}       & \multirow{1}{*}{34.66}        \\ \cline{2-3}
                             & \begin{tabular}[c]{@{}c@{}}KOR\\w/ DeAct. Random Neurons\end{tabular}       & \multirow{1}{*}{34.61}        \\ \cline{2-3}
                             & \begin{tabular}[c]{@{}c@{}}KOR\\w/ DeAct. KOR Neurons in middle layer\end{tabular}       & \multirow{1}{*}{\textit{26.91}}        \\ \cline{2-3}
                             & \begin{tabular}[c]{@{}c@{}}KOR\\w/ DeAct. KOR Neurons in early layer\end{tabular}       & \multirow{1}{*}{\textit{8.80}}        \\ \cline{2-3}
                            \\ \midrule
\multirow{9}{*}{Qwen2.5-7B}  & \multirow{1}{*}{EN}       & \multirow{1}{*}{\textbf{73.38}}        \\ \cline{2-3}
                             & EN Ins.                       & \underline{46.12}         \\ \cline{2-3}
                             & \multirow{1}{*}{KOR}       & \multirow{1}{*}{39.33}        \\ \cline{2-3}
                             & \begin{tabular}[c]{@{}c@{}}KOR\\w/ DeAct. Random Neurons\end{tabular}       & \multirow{1}{*}{38.99}        \\ \cline{2-3}
                             & \begin{tabular}[c]{@{}c@{}}KOR\\w/ DeAct. KOR Neurons in middle layer\end{tabular}       & \multirow{1}{*}{\textit{30.77}}        \\ \cline{2-3}
                             & \begin{tabular}[c]{@{}c@{}}KOR\\w/ DeAct. KOR Neurons in early layer\end{tabular}       & \multirow{1}{*}{\textit{13.85}}        \\ \cline{2-3}
                             \\ \midrule
\multirow{9}{*}{Qwen2.5-32B} & EN                        & \textbf{82.38}                         \\ \cline{2-3}
                             & EN Ins.                       & \underline{75.22}          \\ \cline{2-3}
                             & KOR                       & 68.57                         \\ \cline{2-3}
                             & \begin{tabular}[c]{@{}c@{}}KOR\\w/ DeAct. Random Neurons\end{tabular}       & \multirow{1}{*}{73.97}        \\ \cline{2-3}
                             & \begin{tabular}[c]{@{}c@{}}KOR\\w/ DeAct. KOR Neurons in middle layer\end{tabular}       & \multirow{1}{*}{\textit{64.38}}        \\ \cline{2-3}
                             & \begin{tabular}[c]{@{}c@{}}KOR\\w/ DeAct. KOR Neurons in early layer\end{tabular}       & \multirow{1}{*}{\textit{26.05}}
                             \\ \bottomrule
                             
\end{tabular}
}
\caption{Results of self-correction performance across LLMs. We employ self-correction instructions to elicit the self-correction capability of the models. The full instruction is provided in Appendix~\ref{app:self_correct_instruction}. \textit{EN Ins.} denotes the setting in which LLMs are given English self-correction instructions alongside Korean math problems. For deactivating neurons experiments, we select 100 neurons.}
\label{tab:self_correction_perf}
\end{table}
To measure generation difficulty and compare general tasks with mathematical reasoning and self-correction tasks across English and Korean, we utilize the CAS and DAS metrics (described in Section~\ref{sec:CAS_and_DAS}). Table~\ref{tab:results_CAS_DAS} presents the results of generation difficulty differences between general tasks and mathematical reasoning/self-correction tasks. Similar to perplexity, lower CAS and DAS values indicate that the model experiences less uncertainty when generating the target output.
As shown in Table~\ref{tab:results_CAS_DAS}, the performance gap between English and Korean is substantially larger for mathematical reasoning and self-correction tasks than for general tasks. This suggests that LLMs struggle more with Korean inputs under reasoning-intensive tasks than under general tasks. Moreover, the results indicate that LLMs face considerable difficulty generating self-correction outputs in Korean, whereas generation difficulty is similar for mathematical reasoning and self-correction outputs in English.
We also evaluate self-correction performance by prompting LLMs to revise incorrect solutions and checking whether they ultimately arrive at the correct answer. Table~\ref{tab:self_correction_perf} shows that LLMs generally succeed in self-correcting in English but often fail to do so in Korean.
To further validate this observation, we analyze the activation of Korean-specific neurons \cite{zhaolarge} across different input types: general, mathematical reasoning, and self-correction. The results, illustrated in Figure~\ref{fig:ratio_activated_Koran}, show that the activation ratio of language-specific neurons in the early layers of LLMs is significantly lower for reasoning-intensive inputs compared to general ones. These results imply that the Korean-specific neurons responsible for processing Korean may not be effectively aligned for mathematical reasoning and self-correction, especially compared to their role in general tasks \cite{tang-etal-2024-language, zhaolarge}. We note that the Korean-specific neurons identified in our study are not neurons that directly “think in Korean.” Rather, it is more accurate to view them as neurons that facilitate thinking in English. Thus, they should be interpreted not as ``neurons activated only by Korean,'' but as ``neurons important for handling Korean inputs internally.''
Based on these findings, we manually deactivate Korean-specific neurons\footnote{For 7–8B models, we typically identified 300–500 language-specific neurons. Therefore, we believe that selecting 100 neurons is a reasonable choice, and we report the layer-wise distribution of these neurons in the Appendix~\ref{appendix_neuron_distribution}.} in the early layers and observe a significant performance degradation in both mathematical reasoning and self-correction tasks (Table~\ref{tab:self_correction_perf}). We also compare this with the deactivation of randomly selected neurons in the middle layers—typically associated with English reasoning—and find that deactivating early-layer neurons has a more substantial impact. Motivated by these observations, we fine-tune the Korean-specific neurons in the early layers of LLMs using a self-correction code-switching dataset. This targeted tuning enhances internal translation \cite{zhaolarge} and facilitates a more effective elicitation of the LLMs’ inherent reasoning capabilities.

\subsubsection{Results of Language-Specific Neuron Tuning}
\begin{table*}[t]
\centering
\small
\resizebox{\linewidth}{!}{
\begin{tabular}{llccccc}
\toprule
\multirow{2}{*}{Model}        & \multirow{2}{*}{Method}                                                                     & \multicolumn{5}{c}{Dataset}                                                                 \\ \cline{3-7}
                              &                                                                                             & GSM8K                  & MATH                   & Omni-MATH              & Self-Correction  & General Tasks ($\Delta$)      \\ \midrule
\multirow{7}{*}{Llama3.1-8B}
  & KOR Prompt                                & 57.47 & 31.20 & 11.73 & 34.66 & - \\ \cline{2-7}
  & Continual Pre-training                   & 74.50 & 44.87 & 13.93 & 40.99 & \textbf{+0.2}\\ \cline{2-7}
  & Layer Tuning                             & 73.21 & 43.33 & 12.58 & 35.22 & -3.7 \\ \cline{2-7}
  & LoRA                          & 69.08 & 40.38 & 12.20 & 35.10 & - \\ \cline{2-7}
  & DPO                                      & 73.36 & 44.11 & 13.88 & 35.21 & -2.9 \\ \cline{2-7}
  & Korean-specific Neurons Tuning         & \textbf{74.89} & \textbf{47.25} & \textbf{15.03} & \textbf{43.17} & \textbf{+0.2} \\ \cline{2-7} \\ \cline{2-7}
  & Dataset in EN        & 79.45 & 48.11 & 16.08 & 67.84 & -0.1 \\ \midrule
\multirow{7}{*}{Qwen2.5-7B} 
  & KOR Prompt                            & 66.41 & 50.36 & 18.96 & 39.33 & - \\ \cline{2-7}
  & Continual Pre-training                   & 76.34 & 66.64 & 30.81 & 49.81 & +0.4 \\ \cline{2-7}
  & Layer Tuning                             & 72.82 & 62.91 & 29.02 & 42.70 & -4.4 \\ \cline{2-7}
  & LoRA                           & 70.23 & 61.83 & 28.96 & 40.09 & - \\ \cline{2-7}
  & DPO                                      & 74.50 & 64.13 & 30.55 & 43.86 & -4.2 \\ \cline{2-7}
  & Korean-Specific Neurons Tuning         & \textbf{79.69} & \textbf{68.54} & \textbf{31.19} & \textbf{55.42} & \textbf{+0.9} \\ \cline{2-7} \\ \cline{2-7}
  & Dataset in EN        & 81.35 & 68.87 & 27.29 & 73.38 & +0.1 \\ \bottomrule
\end{tabular}
}
\caption{Results of tuning across all methods. For layer tuning, we select the layer that contains the highest number of language-specific neurons. For language-specific neuron tuning, we choose randomly 100 neurons from the early layers (e.g., below the 12th layer). \textit{Dataset in EN} denotes performance on the English dataset with the CoT prompt. All results are averaged over three runs. For the row labeled ``Dataset in EN'' under \textit{General Tasks}, the baseline corresponds to the model’s performance on English general-task datasets before and after neuron tuning. Our objective is to demonstrate that tuning neurons for Korean mathematical reasoning and self-correction leads to negligible degradation in English performance.}
\label{tab:results_methods}
\end{table*}
To evaluate the effectiveness of various tuning approaches, we assess each approach based on two criteria: (1) whether the method improves performance on mathematical reasoning and self-correction tasks, and (2) whether it preserves the model’s original capabilities in English. For a fair comparison, all methods are trained using our self-correction code-switching dataset. Table~\ref{tab:results_methods} presents the results of all the baselines on both English and Korean datasets. Overall, all methods improve performance on mathematical reasoning and self-correction tasks in Korean, suggesting that the self-correction code-switching data is generally effective for eliciting these capabilities. However, while all methods enhance these capabilities in Korean, only the neuron tuning method achieves performance comparable to that on the original English datasets. These results suggest that tuning Korean-specific neurons in the early layers is particularly effective for eliciting LLMs' inherent reasoning capabilities. Finally, we confirm that neuron-tuning does not degrade the model’s general capabilities in the original language, whereas other methods—except for continual pre-training—significantly impair these abilities. Our findings emphasize that the key to improving specific language reasoning in LLMs is not merely injecting new capabilities, but rather effectively eliciting and aligning the model’s existing reasoning abilities through strategic neuron-level interventions.

\subsubsection{Results of RL After Sufficient Reasoning Ability}
\begin{figure}[t]
    \centering
    \includegraphics[width=\linewidth]{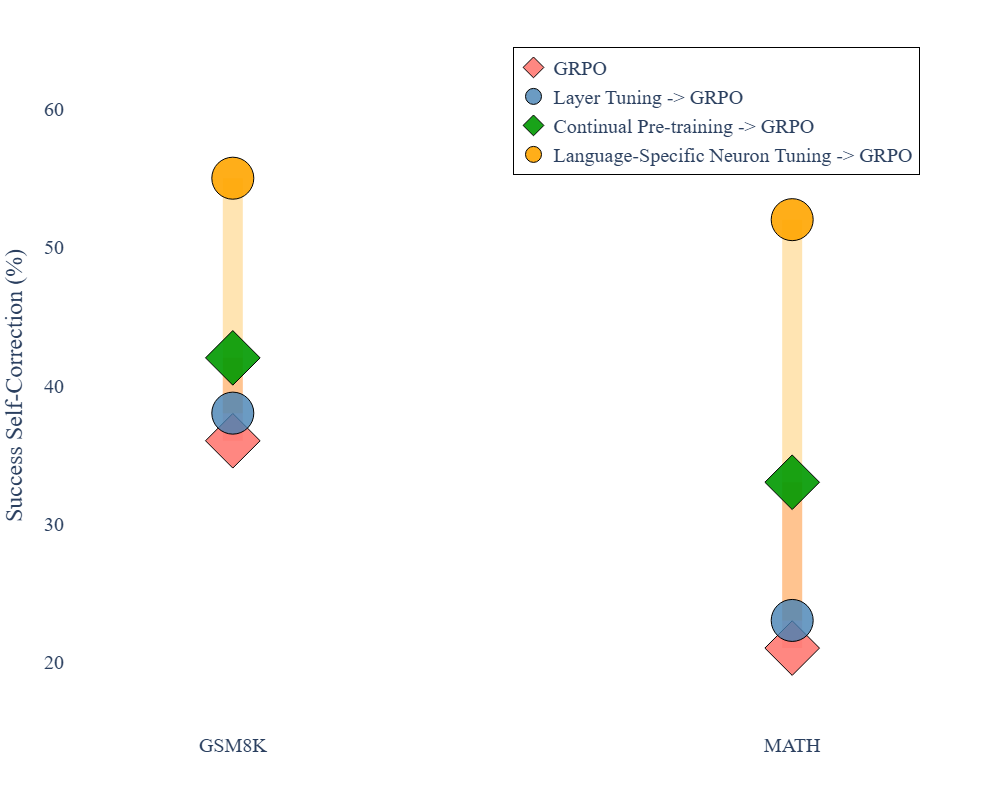}
    \caption{Results of self-correction on Korean GSM8K and MATH datasets across various models after applying GRPO. Llama3.1-8B is used for this experiment. \textit{Successful self-correction} (Y-axis) refers to instances where the model exhibits self-correcting behavior that ultimately leads to the correct answer.}
    \label{fig:analyze_GRPO}
\end{figure}
Figure~\ref{fig:analyze_GRPO} presents the self-correction results on the GSM8K and MATH datasets. We observe that, except for the neuron-tuning method, all baselines exhibit signs of distribution shift on the Korean MATH dataset.
We hypothesize that the generally higher self-correction performance on GSM8K across all methods stems from the similar difficulty levels between the MathDial and GSM8K datasets. As a result, neuron-tuning significantly improved self-correction performance following the application of RL.
We further speculate that the plateaued self-correction performance observed in other methods results from their inability to effectively induce the LLMs’ inherent self-correction capabilities. This may be due to Korean-specific neurons not being sufficiently aligned with the English-centric reasoning pathways that LLMs internally rely upon. Moreover, we re-evaluate the self-correction performance of RL-augmented LLMs after tuning Korean-specific neurons, following the same procedure described in Section~\ref{sec:RL_before_tuning}. Our results demonstrate that applying RL after tuning the early-layer Korean-specific neurons significantly enhances the benefits of RL (Table~\ref{tab:after_train}). This approach successfully induces the self-correction capability of LLMs when presented with mathematical reasoning problems in Korean. For the Qwen3-8B model, the self-correction behavior is minimal on the GSM8K dataset, regardless of whether the Korean-specific neurons are tuned. We attribute this phenomenon to the strong reasoning capabilities of Qwen3-8B: the model can already solve most GSM8K problems without requiring additional correction. As a result, the generated reasoning paths (or trajectories) do not necessarily exhibit revisions or reflective adjustments.

\begin{table*}[t]
\centering
\small
\begin{tabular}{llcc}
\toprule
Model                        & Dataset     & Accuracy ($\Delta$) After RL & Actual Self-Correction Behavior \\
\midrule
\multirow{2}{*}{Llama3.1-8B} & GSM8K (KOR) & +4.3                          & 60\%                             \\
                             & MATH (KOR)  & +2.7                          & 71\%                             \\ \midrule
\multirow{2}{*}{Qwen2.5-7B}  & GSM8K (KOR) & +5.9                          & 53\%                             \\
                             & MATH (KOR)  & +3.2                          & 70\%                            \\ \midrule
\multirow{2}{*}{Qwen3-8B}  & GSM8K (KOR) & +4.0                          & 4\%                             \\
                             & MATH (KOR)  & +8.7                          & 68\%                            \\
\bottomrule
\end{tabular}
\caption{Performance of RL after sufficient Korean reasoning ability.} \label{tab:after_train}
\end{table*}



%% file: sections/analysis.tex
\section{Analysis}
\subsection{Various Settings of Self-Correction Data}
\begin{figure}[t]
    \centering
    \includegraphics[width=1\linewidth]{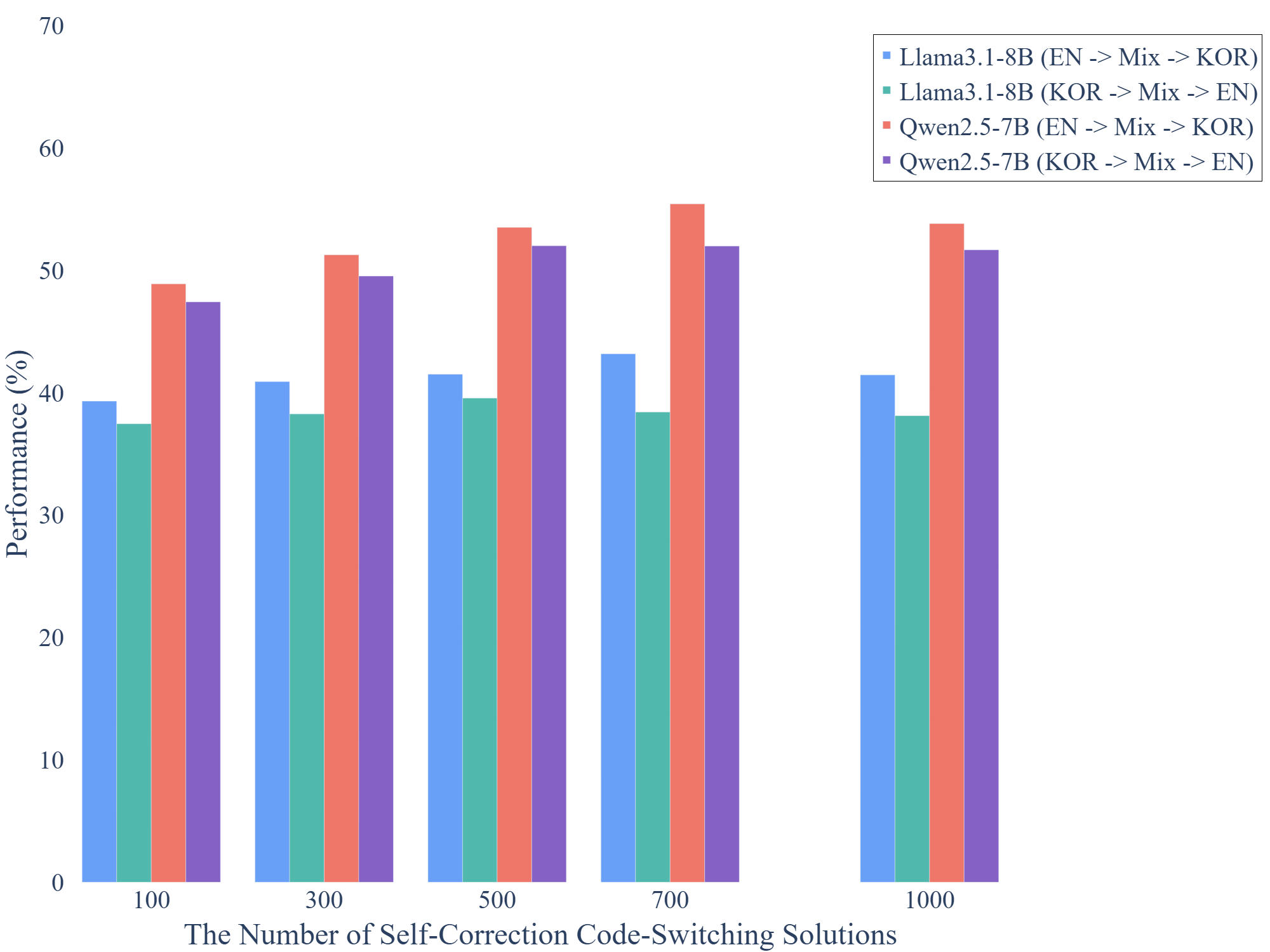}
    \caption{Self-correction performance change according to the number of self-correction code-switching solutions used to fine-tune Korean-specific neurons.}
    \label{fig:analyze_solution_numbers}
\end{figure}
To further analyze the effectiveness of language-specific neuron tuning with self-correction code-switching learning, we conduct empirical experiments by varying the setting of our self-correction code-switching data. 
Here, we analyze the effect of the number of self-correction code-switching solutions used to fine-tune language-specific neurons. We generate 1K solutions and incrementally increase the number of samples used for tuning, analyzing the corresponding changes in self-correction performance. The results are shown in Figure~\ref{fig:analyze_solution_numbers}.
As shown in the figure, performance generally improves with the increased number of self-correction samples. However, a slight performance degradation is observed when tuning with all 1K examples. We speculate that this decline may be due to overfitting, as the number of language-specific neurons is extremely small relative to the total number of parameters in the LLM, making the model more prone to overfitting to the limited dataset.
In addition, we analyze the effect of changing the language sequence in the self-correction code-switching dataset. For constructing this dataset, we consider two generation options for the solution: (1) English → Mixed → Korean, and (2) Korean → Mixed → English. As shown in Figure~\ref{fig:analyze_solution_numbers}, the first option yields significantly better performance than the second. We interpret this as evidence that the English → Mixed → Korean curriculum gives the model an explicit alignment signal, effectively showing ``what to translate to,'' and thereby enabling early-layer neurons to learn a stronger internal translation mechanism. We also hypothesize that when LLMs—even highly capable ones such as GPT-4.1—are prompted to generate Korean before English, the quality of the synthetic data may be degraded. Similar findings have been reported in studies on prompt engineering for English and Korean \cite{ko2025understand}.

%% file: sections/conclusion.tex
\section{Conclusions}
In this study, we investigate whether reinforcement learning (RL) can enhance Korean reasoning capabilities in large language models (LLMs) and under what conditions such improvements are most effective. Our experiments reveal that applying RL in isolation, without sufficient underlying Korean reasoning abilities, leads to limited gains—highlighting a significant disparity compared to English. We show that RL becomes markedly more effective when LLMs are first fine-tuned to possess strong Korean reasoning capabilities, particularly through tuning language-specific neurons that enhance internal translation. Our findings emphasize that the key to improving multilingual reasoning in LLMs is not merely injecting new capabilities in low-resource languages, but rather effectively eliciting and aligning the model’s existing reasoning abilities through strategic neuron-level interventions.

%% file: sections/Limitations.tex
\section*{Limitations}
While our study demonstrates that aligning language-specific neurons can significantly enhance Korean reasoning and self-correction capabilities in LLMs, several limitations remain. First, our approach primarily focuses on tuning neurons in early layers based on activation patterns from mathematical reasoning and self-correction tasks. This might not generalize well to other domains (e.g., commonsense or dialogue reasoning) or languages beyond Korean. Second, although our method effectively improves alignment and benefits reinforcement learning (RL), it does not directly enhance the model’s native Korean understanding or generation capabilities, such as fluency or cultural nuance. Third, our experiments rely on relatively small-scale code-switching data, and the scalability and robustness of our approach in truly low-resource or zero-resource scenarios remain unexplored. Lastly, while we verify our findings on select LLM architectures (e.g., LLaMA and Qwen), further validation is needed across a broader range of model families and sizes to assess the universality of neuron-level interventions.

%% file: sections/Acknowledgement.tex
\section*{Acknowledgement}
This work was supported by the Institute of Information \& Communications Technology Planning \& Evaluation (IITP) grant funded by the Korea Government (MSIT) (No. RS-2023-00216011, Development of Artificial Complex Intelligence for Conceptually Understanding and Inferring like Human) and IITP grant funded by the Korea Government (MSIT) (No. RS-2024-00338140, Development of learning and utilization technology to reflect sustainability of generative language models and up-to-dateness over time)

%% file: sections/appendix.tex
\appendix
\input{sections/related_work}
\section{Details for Self-Correction} \label{app:self-correction_details}
\begin{table*}[t]
\resizebox{\linewidth}{!}{
\begin{tabular}{l|p{0.9\linewidth}}
\toprule
\textbf{Category} & \textbf{Content} \\
\hline
\textbf{Problem (EN+KOR)} & Bruno wants to buy two and one-half dozens of pens. How many pens will he have? (Bruno는 2와 1/2 다스의 펜을 사고 싶어 합니다. 그는 몇 개의 펜을 가지게 될까요?) \\
\hline
\textbf{Self-Correction (EN)} & One dozen is equal to 12, so two dozens is equal to 2 x 12 = 24 pens. ... Thus, two and one-half dozens of pens is equal to 24 + 18 = 42 pens.  
\textbf{Wait... something doesn't feel right. I'm calculating "two and one-half dozens," not "two dozens plus one and a half dozen."  
Let me take a step back and rethink what "two and one-half dozens" actually means.  
"Two and one-half dozens" means 2.5 dozens.  
Since one dozen equals 12 pens, I should multiply:  
2.5 × 12 = 30 pens.} \\
\hline
\textbf{Self-Correction (KOR)} & 한 다스는 12개이므로, 두 다스는 2 x 12 = 24개입니다. ... 따라서 2와 1/2 다스는 24 + 18 = 42개입니다.  
\textbf{잠깐만요... 뭔가 이상해요. 지금은 "2와 1/2 다스"를 계산해야 하는데, 나는 "2 다스 + 1과 1/2 다스"를 더해버렸어요.  
조금만 다시 생각해볼게요.  
"2와 1/2 다스"는 소수로 쓰면 2.5 다스예요.  
한 다스가 12개니까, 2.5 x 12 = 30개예요.} \\
\hline
\textbf{Self-Correction (Code-Switching)} & \textbf{Wait… 뭔가 이상한데, I think I just added 2 dozens and 1.5 dozens separately, which isn’t what the question asked.  
"two and one-half dozens"는 그냥 2.5 dozens인데, I treated it like 2 + 1.5 = 3.5 dozens without realizing.  
That was totally my mistake — 계산은 맞았지만 interpretation이 완전히 틀렸어.  
So instead of doing 24 + 18, I should’ve just done 2.5 × 12.  
한 dozen이 12개니까, 2.5 dozens면 2.5 × 12 = 30 pens가 정답이지.} \\ \bottomrule
\end{tabular}
}
\caption{Examples of self-correction and code-switching sample.} \label{tab:data_example}
\end{table*}

\paragraph{Generating Instruction} The instruction to generate a self-corrected solution is as follows:
\begin{instructionsbox}[Self-Correction Generating Instruction]
\begin{lstlisting}
Given a problem and its incorrect solution, correct solution through backtracking (e.g., self-correction or reflection). [Additionally, ensure that the response should be Korean language.]
Problem: {problem}
Incorrect Solution until first error: {correct_solution} + [However, but, wait]
Corrected Solution: {Model Output}
\end{lstlisting}
\end{instructionsbox}
Table~\ref{tab:data_example} presents examples of the generated self-correction data in both English and Korean.

\paragraph{Instruction for Measuring CAS and DAS} The instruction to measure the CAS and DAS is as follows:
\begin{instructionsbox}[Measuring CAS and DAS Instruction]
\begin{lstlisting}
Given a problem and its ground-truth solution, generate a new solution that initially contains errors but ultimately arrives at the correct solution through backtracking (e.g., self-correction or reflection). [Additionally, ensure that the response should be Korean language.]
Problem: {problem}
Ground-truth Solution: {Self-Corrected Solution}
\end{lstlisting}
\end{instructionsbox}
We measure CAS and DAS for only \{Self-Corrected Solution\}.

\section{Detailed Description for Identifying Language-Specific Neurons} \label{app:identifying_neurons}
\subsection{Parallel Neuron Detection}

While sequential neuron detection involves iteratively computing the importance of each neuron for every input—making the process computationally expensive—we propose a parallel approach to accelerate the detection.

\paragraph{Feed-Forward Network (FFN).} 
In modern open-source LLMs, the feed-forward network (FFN) in a Transformer layer is typically formulated as:
\begin{equation}
\text{FFN}(x) = \left( \text{SiLU}(W_{\text{gate}} x) \cdot W_{\text{up}} x \right) W_{\text{down}},
\end{equation}
where $x \in \mathbb{R}^{l \times d_{\text{model}}}$ is the input embedding, $W_{\text{gate}}, W_{\text{up}} \in \mathbb{R}^{d_{\text{model}} \times d_{\text{inter}}}$, and $W_{\text{down}} \in \mathbb{R}^{d_{\text{inter}} \times d_{\text{model}}}$. The importance of the $k$-th neuron in $W_{\text{up}}$ can be efficiently computed as:
\begin{equation}
\text{Imp}(W_{\text{up}}[:, k] \mid c) = \left\| (h_{\text{ffn}} \cdot \text{Mask}[k]) W_{\text{down}} \right\|_2,
\end{equation}
where $h_{\text{ffn}}$ is the intermediate activation before $W_{\text{down}}$ and $\text{Mask}[k]$ is a one-hot vector with 1 at the $k$-th index. By stacking one-hot vectors as a diagonal matrix $\text{Mask}$, we parallelize the importance computation as:
\begin{equation}
\text{Imp}(W_{\text{up}} \mid c) = \left\| (h_{\text{ffn}} \cdot \text{Mask}) W_{\text{down}} \right\|_2.
\end{equation}
This formulation also allows us to equivalently compute the importance of neurons in $W_{\text{down}}$ by leveraging their symmetry with $W_{\text{up}}$.

\paragraph{Self-Attention Network.}
The self-attention mechanism for input $x$ is defined as:
\begin{equation}
\small
\text{Attention}(x) = \text{Softmax} \left( \frac{W_Q x \cdot (W_K x)^T}{\sqrt{d}} \right) W_V x,
\end{equation}
where $W_Q, W_K, W_V \in \mathbb{R}^{d_{\text{model}} \times d_{\text{mid}}}$. Since $W_V$ is outside the softmax computation, its importance can be estimated using the FFN-style equation.

For $W_Q$, we estimate the importance of its $k$-th neuron by measuring the change in attention weights after zeroing out that neuron:
\begin{equation}
\centering
\small
\Delta_k(x) = W_Q(x)[:, k] \cdot W_K(x)[k, :] \in \mathbb{R}^{l \times l}.
\end{equation}
Then, we compute the importance as the change in output caused by this attention shift:
\begin{equation}
\centering
\tiny
\begin{split} 
    & \text{Imp}(W_Q[:, k] \mid c) \approx \\ & \left\|  \text{Softmax} \left( \frac{W_Q x \cdot W_K^T x - \Delta_k(x)}{\sqrt{d}} \right) -  \text{Softmax} \left( \frac{W_Q x \cdot W_K^T x}{\sqrt{d}} \right) \right\|_2
\end{split}
\end{equation}

To accelerate computation, this can also be parallelized by constructing the full $\Delta(x)$ tensor as:
\begin{equation}
\centering
\small
\begin{split}
    \Delta(x) =  & W_Q(x).\text{reshape}(l, 1, d_{\text{mid}}) \ \cdot \\ & W_K(x). \text{reshape}(1, l, d_{\text{mid}}) \\ & \in \mathbb{R}^{l \times l \times d_{\text{mid}}},
\end{split}
\end{equation}
allowing:
\begin{equation}
\centering
\tiny
\begin{split}
    & \text{Imp}(W_Q \mid c) \approx \\ & \left\| \text{Softmax} \left( \frac{W_Q x \cdot W_K^T x - \Delta(x)}{\sqrt{d}} \right) - \text{Softmax} \left( \frac{W_Q x \cdot W_K^T x}{\sqrt{d}} \right) \right\|_2
\end{split}
\end{equation}
A similar approach is used to compute $\text{Imp}(W_K \mid c)$ due to its symmetry with $W_Q$.

\subsection{Korean-specific Neuron Distribution} \label{appendix_neuron_distribution}
\begin{figure}
    \centering
    \includegraphics[width=1.0\linewidth]{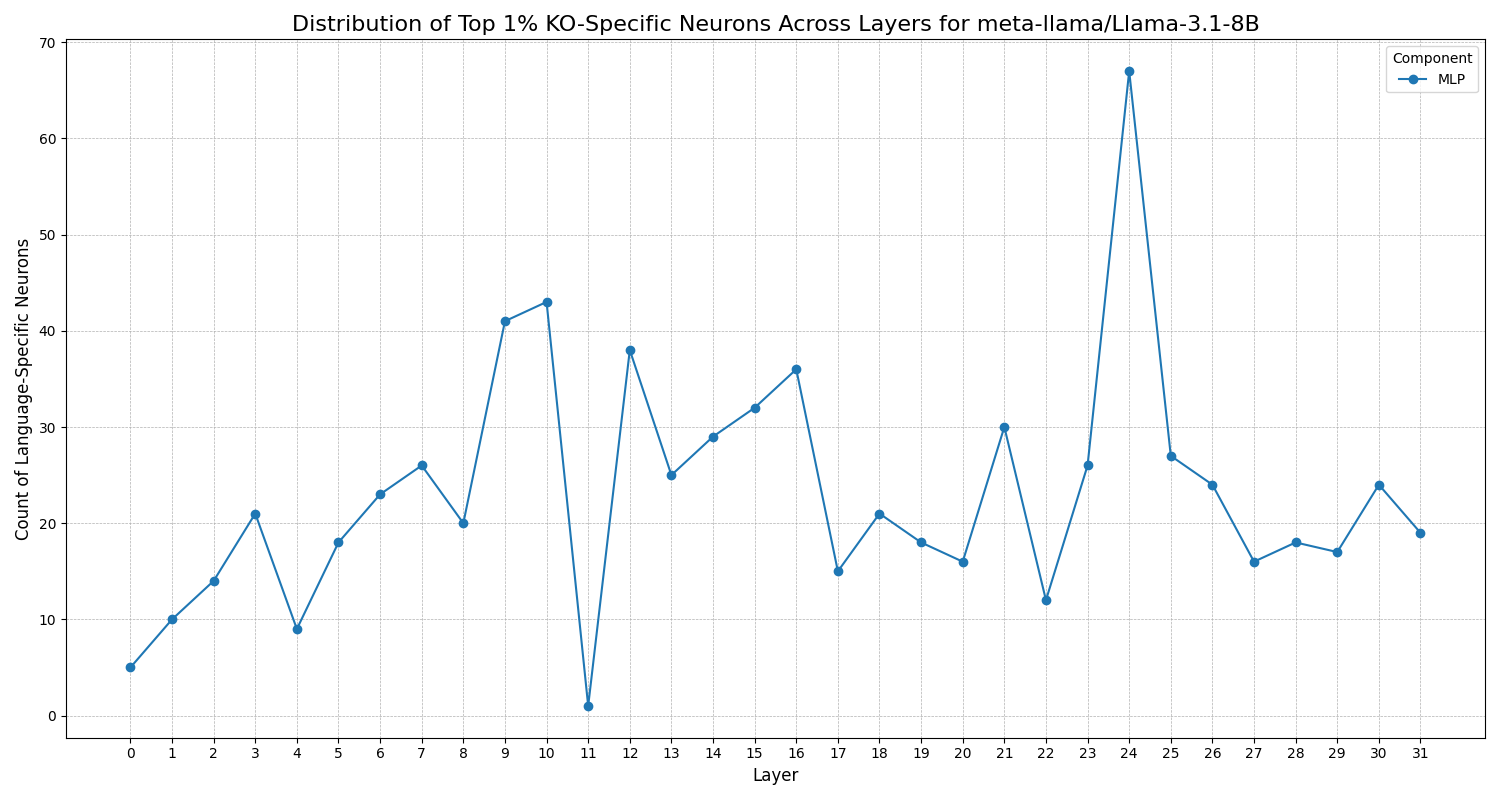}
    \caption{Distribution of Korean-specific neurons in the Llama3.1-8B MLP modules.}
    \label{fig:neuron_distribution_llama3.1}
\end{figure}

\begin{figure}
    \centering
    \includegraphics[width=1.0\linewidth]{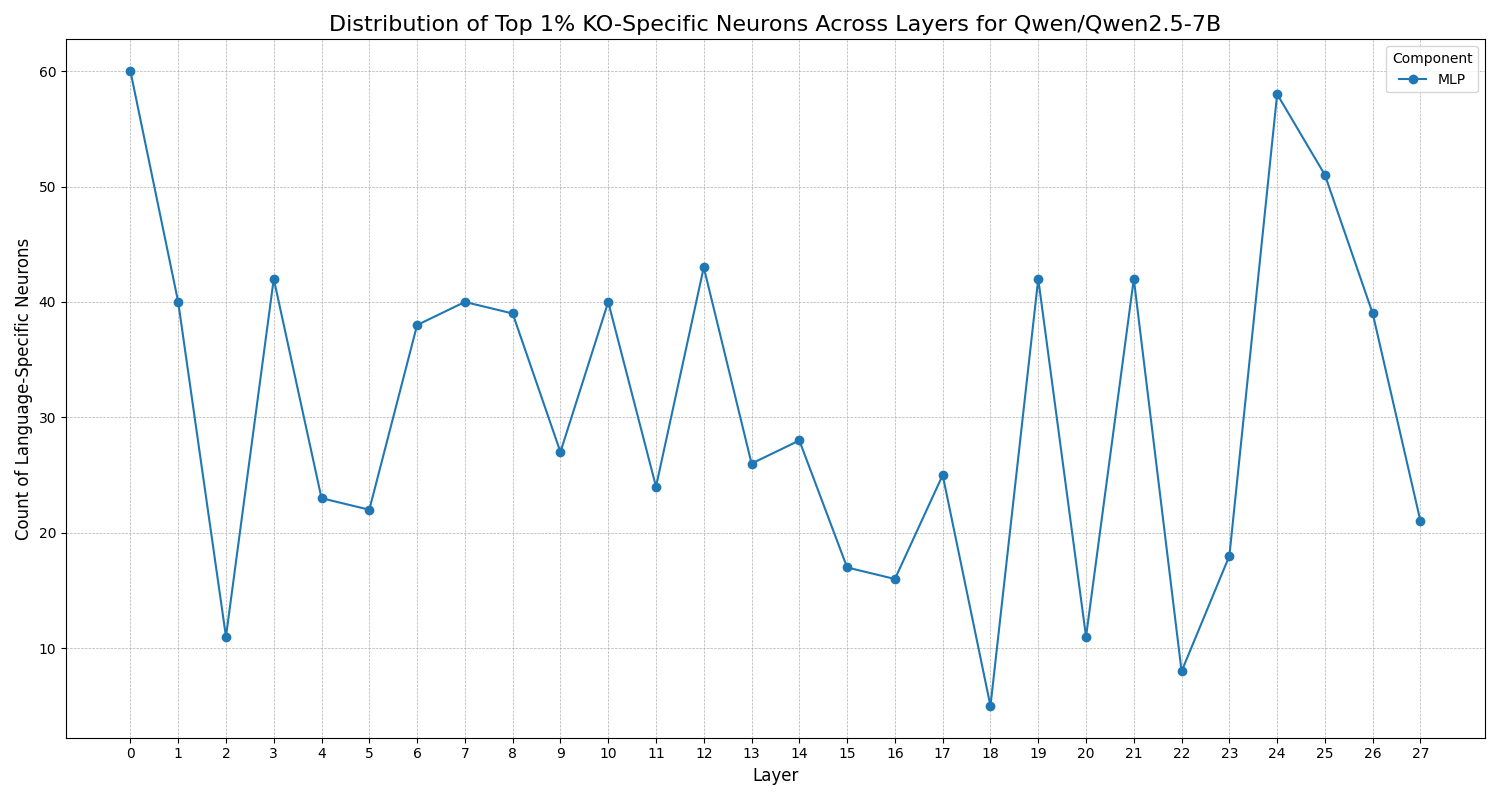}
    \caption{Distribution of Korean-specific neurons in the Qwen2.5-7B MLP modules.}
    \label{fig:neuron_distribution_qwen2.5}
\end{figure}

\begin{figure}
    \centering
    \includegraphics[width=1.0\linewidth]{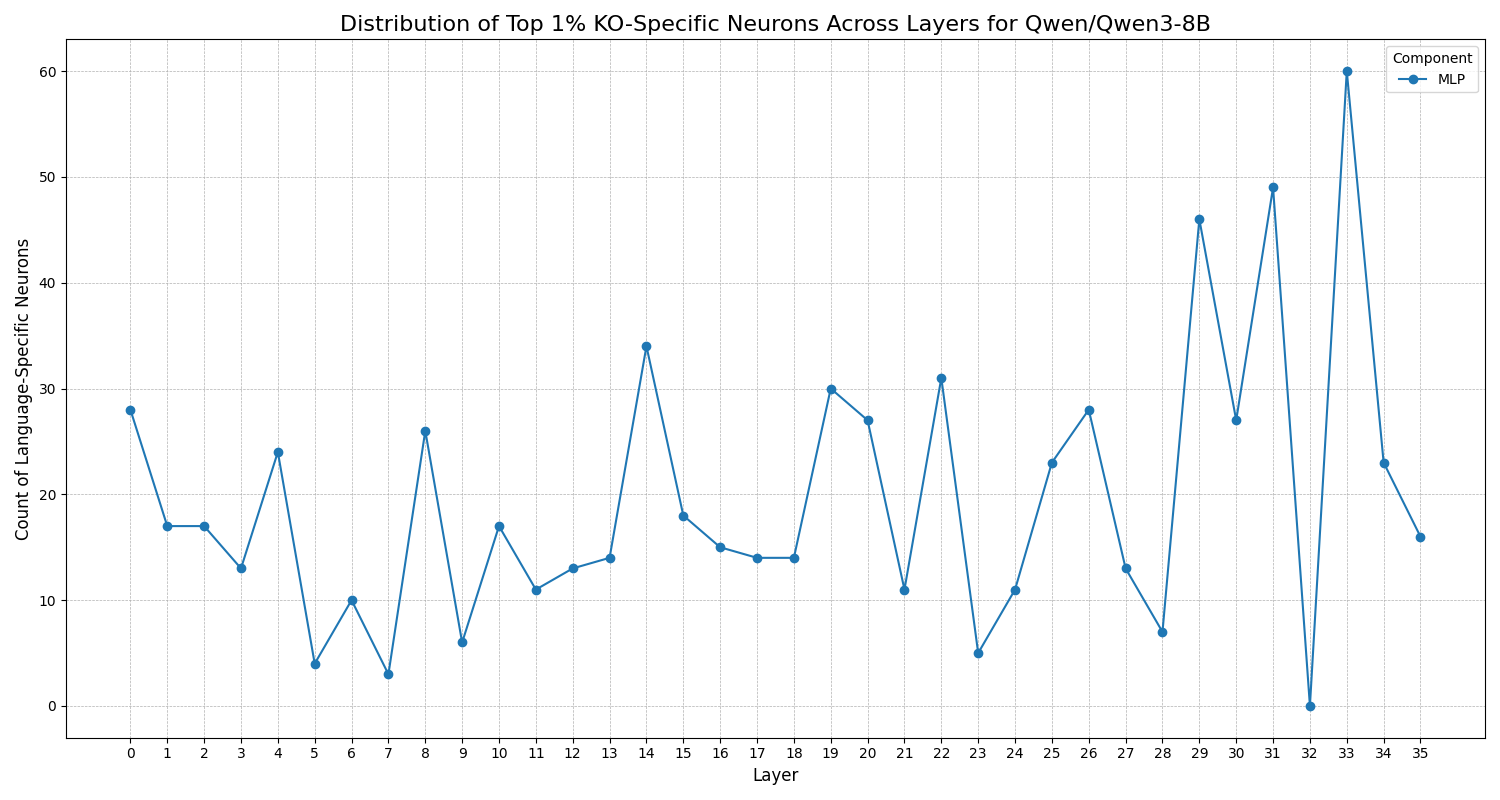}
    \caption{Distribution of Korean-specific neurons in the Qwen3-8B MLP modules.}
    \label{fig:neuron_distribution_qwen3}
\end{figure}

Figures~\ref{fig:neuron_distribution_llama3.1}, \ref{fig:neuron_distribution_qwen2.5}, and \ref{fig:neuron_distribution_qwen3} show the distribution of Korean-specific neurons in the MLP modules of the Llama3.1, Qwen2.5, and Qwen3 models.

\section{Experimental Settings} \label{app:experimental_setting_details}
The temperature setting was fixed at 0.0 (i.e., greedy decoding), with both top-p and top-n configured to 1. However, for a reasoning optimized model such as Qwen3, we used the optimal setting for reasoning mode. All experiments were conducted using 8 NVIDIA A100 80GB GPUs.
\paragraph{LoRA}We set the LoRa alpha to 64 and dropout to 0.1. We set r, which is the dimension size of Lora, to 128. We set the batch size to 12 and the accumulation steps to 5. We also set the learning rate to 2e-4.\footnote{Our code and all instructions will be available on Github}
\paragraph{Continual Pre-training} For continual pre-training, we conduct training on 8 A100 GPUs over 3 epochs, using a context length of 4,096 tokens and a warm-up ratio of 0.01. The optimizer applies a weight decay of 0.01, and the peak learning rate is set to 2e-5, following an inverse square root decay schedule. Training is performed in FP16 precision using DeepSpeed and FlashAttention.

\paragraph{DPO} For DPO training, we construct preference pairs such that the model is encouraged to prefer self-corrected solutions over ground-truth solutions. The ground-truth solutions solve the math problem directly, while the self-corrected ones include explicit reflective steps (e.g., identifying errors, “aha” moments). Both types of solutions are structured with the same code-switching format (English → mixed → Korean), ensuring consistency in training. All model parameters were updated during training.

\paragraph{GRPO}
We adopt GRPO as our RL algorithm. During training, we use a batch size of 32, while validation is performed with a batch size of 256. The learning rate is set to $3 \times 10^{-7}$. To improve memory efficiency, the update mini-batch size is set to 8. The KL divergence loss coefficient is 0.001. For each prompt, 8 candidate responses are generated to compute relative rewards. Training is conducted for a single epoch, with evaluations performed every 10 steps. The maximum response length is limited to 4,096 tokens. The model takes the previous 12 utterances as input context. For training, we use a self-correction dataset in Korean. All experiments are conducted on a system equipped with 8 NVIDIA A100 80GB GPUs.

\section{Statistic of Datasets} \label{app:statistics}
Table \ref{tab:HRM8K_detail} shows the statistics and description of the HRM8K dataset. We used the prior set of HRM8K for our experiments.
\begin{table*}[t]
\centering
\resizebox{\linewidth}{!}{
\begin{tabular}{llcp{8cm}l}
\toprule
\textbf{Category} & \textbf{Subset} & \textbf{\# of Instances} & \textbf{Short Description} \\
\midrule
\multirow{10}{*}{KSM: \textbf{1.4K Total}} 
  & KMO & 730 & Mathematics competition for high school students in South Korea; top performers are selected as representatives for the IMO. \\
  & KJMO & 62 & Junior division of the KMO, intended for students up to age 13. \\
  & CSAT & 210 & Questions from the Korean national university entrance exam and official mock exams; we include only questions with an error rate exceeding 70\%. \\
  & KMS & 82 & University-level math olympiad organized by the Korean Mathematical Society. \\
  & TQ & 344 & Questions from the national assessment test for math teacher certification. \\
\midrule
\multirow{8}{*}{Prior Sets: \textbf{6.5K Total}} 
  & GSM8K & 1,319 & Grade school math word problems written by human problem authors. \\
  & MATH & 2,885 & Competition-level mathematics problems with numeric answers only. \\
  & Omni-MATH & 1,909 & Olympiad-level problems collected from international and Chinese math competitions; only questions with numeric answers are included. \\
  & MMMLU & 470 & A subset of the MMLU dataset translated by professional human translators. \\
\bottomrule
\end{tabular}
}
\caption{Details of the HRM8K dataset.}
\label{tab:HRM8K_detail}
\end{table*}

\section{Self-Correction Instructions} \label{app:self_correct_instruction}
\begin{instructionsbox}[Self-Correction Instruction]
\begin{lstlisting}
Please solve the given problem.
Make sure to carefully check for any errors in the process of solving it.
\end{lstlisting}
\end{instructionsbox}



%% file: sections/related_work.tex
\section{Related Work}
\subsection{Multilingual LLMs}
Due to the imbalance of pre-training data across languages, LLMs tend to exhibit inferior performance in low-resource languages. To mitigate this gap, recent studies have proposed effective language transfer methods, including vocabulary extension, continual pre-training \cite{yoo2024code, han2025trillion7btechnicalreport}, and adapter tuning \cite{shen-etal-2024-language, jiang2025franken}. For example, \citet{han2025trillion7btechnicalreport} introduced Trillion 7B, a Korean-centric multilingual LLM, by applying tailored vocabulary construction and cross-lingual document attention (XLDA) to effectively manage cross-lingual interactions. In contrast to these approaches, this study identifies language-specific neurons—particularly those related to Korean—and fine-tunes them to enhance mathematical reasoning and self-correction capabilities.

\subsection{Model Interpretability}
Recently, many studies have investigated the relationship between the internal mechanisms of LLMs and their specific capabilities \cite{geva2021transformer, hou-etal-2023-towards, yang2024large, yu-ananiadou-2024-large, yu2024interpreting, yu-ananiadou-2024-neuron, fan-etal-2025-slam, xu-etal-2025-lets}. For instance, \citet{yu2024interpreting} analyzed attention heads and feed-forward network (FFN) neurons to identify the specific parameters responsible for arithmetic ability, observing that only a small number of heads significantly impact arithmetic tasks.
Although these interpretability studies have provided valuable insights into the inner workings of LLMs, most focus primarily on English. In the context of multilingual interpretability, \citet{tang-etal-2024-language} introduced Language Activation Probability Entropy (LAPE) to identify language-specific neurons, showing that particular languages are predominantly processed by a small subset of neurons located in the top and bottom layers of the model. Similarly, \citet{zhaolarge} proposed Parallel Language-Specific Neuron Detection (PLND) to investigate how LLMs process multilingual input. Their results confirmed that LLMs tend to internally translate non-English inputs into English using neurons in the bottom layers, perform reasoning in English in the middle layers, and generate output in the original input language using neurons in the top layers. Moreover, they demonstrated that deactivating a small subset of language-specific neurons leads to a significant drop in the model's performance for the corresponding language \cite{tang-etal-2024-language, zhaolarge}.
These neuron-level studies share a common observation: a relatively small subset of neurons plays a dominant role in processing specific abilities or languages. Furthermore, it is evident that LLMs primarily perform reasoning in English within their middle layers. This is supported by the finding from \citet{tang-etal-2024-language} that language-specific neurons are concentrated in the top and bottom layers, and by \citet{zhaolarge}'s observation that early layers are responsible for internal translation and top layers for language-specific output generation.
Building on these findings, we identify Korean-specific neurons in the early layers of LLMs using the existing method for detecting language-specific neurons. We further observe that while these neurons are highly activated for general tasks, they are notably less active when processing inputs involving mathematical reasoning and self-correction. We hypothesize that these neurons—responsible for internally translating Korean into English—are not yet aligned to support reasoning and self-correction capabilities. Our empirical results demonstrate that aligning these neurons to improve internal translation yields significantly better reasoning performance in Korean compared to enabling such capabilities through continual pre-training alone.
